\documentclass[twoside]{article}
\usepackage{ecj,palatino,epsfig,latexsym,natbib}

\usepackage[ruled,vlined,linesnumbered]{algorithm2e}
\usepackage{textcomp}
\usepackage{stfloats}
\usepackage{url}
\usepackage{verbatim}
\usepackage{graphicx}
\usepackage {makecell}
\usepackage{multirow}
\usepackage{amsmath,amssymb,amsfonts}
\usepackage{pdfpages}
\usepackage{float}
\usepackage{tikz,xcolor,hyperref}

\definecolor{lime}{HTML}{A6CE39}
\DeclareRobustCommand{\orcidicon}{
    \hspace{-2.5mm}
    \begin{tikzpicture}
    \draw[lime, fill=lime] (0,0)
    circle [radius=0.16]
    node[white] {{\fontfamily{qag}\selectfont \tiny ID}};
    \draw[white, fill=white] (-0.0625,0.095)
    circle [radius=0.007];
    \end{tikzpicture}
    \hspace{-2mm}
}

\foreach \x in {A, ..., Z}{%
    \expandafter\xdef\csname orcid\x\endcsname{\noexpand\href{https://orcid.org/\csname orcidauthor\x\endcsname}{\noexpand\orcidicon}}
}

\begin{document}

\ecjHeader{x}{x}{xxx-xxx}{201X}{Multi-Representation Genetic Programming}{Z. Huang, Y. Mei, F. Zhang, M. Zhang, and W. Banzhaf}
\title{\bf Multi-Representation Genetic Programming: A Case Study on Tree-based and Linear Representations}  

\author{\name{\bf Zhixing Huang\orcidA{}}  \hfill 
\addr{zhixing.huang@ecs.vuw.ac.nz}\\ 
\name{\bf Yi Mei\orcidC{}}  \hfill 
\addr{yi.mei@ecs.vuw.ac.nz}\\ 
\name{\bf Fangfang Zhang\orcidB{}}  \hfill 
\addr{fangfang.zhang@ecs.vuw.ac.nz}\\ 
\name{\bf Mengjie Zhang\orcidD{}}  \hfill 
\addr{mengjie.zhang@ecs.vuw.ac.nz}\\ 
        \addr{Centre for Data Science and Artificial Intelligence \& School of Engineering and Computer Science, Victoria University of Wellington, Wellington, 6140, New Zealand}
\AND
       \name{\bf Wolfgang Banzhaf\orcidE{}} \hfill \addr{banzhafw@msu.edu}\\
        \addr{Department of Computer Science and Engineering, BEACON Center for the Study of Evolution in Action, and Ecology, Evolution and Behavior Program, Michigan State University, East Lansing, MI 48864, USA}
}

\maketitle

\begin{abstract}

 Existing genetic programming (GP) methods are typically designed based on a certain representation, such as tree-based or linear representations. These representations show various pros and cons in different domains. However, due to the complicated relationships among representation and fitness landscapes of GP, it is hard to intuitively determine which GP representation is the most suitable for solving a certain problem.
Evolving programs (or models) with multiple representations simultaneously can alternatively search on different fitness landscapes since representations are highly related to the search space that essentially defines the fitness landscape.
Fully using the latent synergies among different GP individual representations might be helpful for GP to search for better solutions.
However, existing GP literature rarely investigates the simultaneous effective use of evolving multiple representations. To fill this gap, this paper proposes a multi-representation GP algorithm based on tree-based and linear representations, which are two commonly used GP representations. In addition, we develop a new cross-representation crossover operator to harness the interplay between tree-based and linear representations. Empirical results show that navigating the learned knowledge between basic tree-based and linear representations successfully improves the effectiveness of GP with solely tree-based or linear representation in solving symbolic regression and dynamic job shop scheduling problems.

\end{abstract}


\section{Introduction}
Genetic programming (GP) has shown impressive performance in many machine learning domains such as classification \cite{Dimmy2023,Bi2022b} and symbolic regression \cite{David2023,Huang2022SLGP}.
Over the years, many GP variants have been proposed, such as tree-based GP (TGP)\cite{Koza1992}, Linear GP (LGP) \cite{Nordin1994,Nordin1997}, Cartesian GP \cite{Miller1999}, gene expression programming \cite{Ferreira2001}, graph-based genetic programming \cite{Atkinson2018}, and grammar-guided GP \cite{Forstenlechner2017}.
Traditionally, GP only evolves individuals within a single representation. The individual representation (and its corresponding search mechanism) directly defines the search space and the corresponding fitness landscape.

Generally speaking, a GP representation is expected to be suitable for only a subset of problems. Although some existing studies have investigated the performance of different GP representations in solving different problems based on empirical comparisons \cite{Wilson2008,Sotto2021Graph}, extending such kind of knowledge to unseen domains is difficult, and such investigations are often too time-consuming and it is hard to cover all different branches and variants of a problem. 
When encountering an emerging application or a new problem, users have scarce domain knowledge in selecting a GP representation.
To make full use of different GP representations and enhance the search performance of existing GP methods, it would be interesting to investigate whether the different GP representations can cooperate in solving a single task.

To this end, this paper proposes a new Multi-Representation GP (MRGP) algorithm that simultaneously evolves individuals with more than one representation. This paper focuses on the MRGP with two typical GP representations, tree-based (i.e., TGP) \cite{Koza1992} and linear-based (i.e., LGP) \cite{Brameier2007} representations, denoted as MRGP-TL.
TGP and LGP have very different program representations, TGP with tree-based structures, and LGP with instruction lists. Consequently, the structures of TGP programs are usually wide, while the topological structures of LGP programs are usually long and narrow \cite{Huang2023MLSI}.
It is likely that the two GP methods should have different fitness landscapes and search biases.
MRGP-TL simultaneously evolves sub-populations with tree-based and linear GP representations for a single task and exchanges search information across representations, aiming to obtain more diverse useful building blocks with a higher chance to find better solutions.


The major challenge of designing MRGP is that different GP representations cannot be exchanged building blocks directly. For example, tree-based representations in TGP do not contain information about registers in LGP.
Furthermore, even within the LGP framework, individuals with different maximum numbers of registers cannot exchange instruction segments directly since it likely produces instructions with invalid registers.
In addition, instruction outputs (register values) in LGP individuals can be reused by more than one subsequent instruction because of the graph-based structure (LGP individuals can be decoded into a graph, see Section \ref{sec:diffGPrep}), but tree nodes in standard TGP can only be used once. Thus, directly exchanging building blocks from different representations might not always produce valid offspring.

To address the above issue, here we propose to unify the building blocks of different GP representations into \emph{adjacency lists} to effectively exchange building blocks between tree-based and linear GP representations. An adjacency list is a common and universal representation of graphs and can represent different kinds of topological structures. 
Besides, adjacency lists can convey graph information such as the frequency of different nodes and their connections. Although adjacency lists can be an intermediate representation for tree-based and linear representations, this does not mean that an adjacency list is a more effective GP representation than tree-based or linear representations because of the two following reasons. 
First, existing literature shows that evolving computer programs based on graph-based structures is not always better than tree-based and linear representations \cite{Sotto2021Graph}. Different representations have their own pros and cons for different tasks.
Second, a conventional adjacency list relies on graph node indices to distinguish graph nodes. But different graphs (i.e., GP individuals) likely have different indices for the same building blocks (e.g., a three-node building block ``$x_1 + x_2$'' might be indexed as ``$\mathcal{A} \rightarrow[\mathcal{B}, \mathcal{C}]$'' and ``$\mathcal{D} \rightarrow[\mathcal{E}, \mathcal{F}]$'' in two different GP individuals). The indexing mechanisms for adjacency list representations might be too complicated to show obvious advantages.

This paper has three main contributions:
\begin{enumerate}
  \item We propose a multi-representation evolutionary framework for GP methods. The proposed evolutionary framework simultaneously evolves multiple sub-populations, each with a distinct GP representation and all solving the same task. By sharing the building blocks among representations, the multi-representation evolutionary framework stimulates GP to search for more effective solutions. To the best of our knowledge, this paper is the first to highlight that cooperation among different GP representations is beneficial for GP search performance.
 \item To implement the new evolutionary framework, we propose an MRGP algorithm based on two representative GP systems (i.e., TGP and LGP), denoted as MRGP-TL. The newly proposed algorithm simultaneously evolves two sub-populations, one with tree-based representation and the other with linear representation. A new crossover operator is introduced based on the adjacency list to exchange building blocks from tree-based and linear subpopulations in the course of evolution. Since many other GP representations can also be seen as graphs and described by adjacency lists, the newly proposed crossover between tree-based and linear representations can inspire future work in other GP representations. 

 \item This paper verifies the effectiveness of MRGP-TL by two substantially different applications. The two applications are symbolic regression and automatic design of decision rules in dynamic job shop scheduling problems, which cover a wide range of applications \cite{Cai2020,Song2019,DAriano2015,Schauer2013}.
The results show that simultaneously evolving basic tree-based and linear representations is more effective than the original single-representation methods in both problem domains. Furthermore, by extending the multi-representation evolutionary framework to other state-of-the-art methods, we got significant performance improvement for dynamic job shop scheduling problems.
 


\end{enumerate}

\section{Literature Review}
\subsection{Tree-based and Linear GP Representations}\label{sec:diffGPrep}
TGP \cite{Koza1992} uses tree-based representation, where each individual encodes a computer program as a tree. Every tree node represents a function or a terminal (i.e., input feature). Function nodes accept inputs from their sub-trees and deliver results to their parent nodes. Each tree node has up to one parent node.
All intermediate results from sub-trees are aggregated at the root, with the root outputting the final result of the program. 
Tree-based representation has been successfully applied to different domains such as classification \cite{devarriya2020unbalanced,Bi2022}, symbolic regression \cite{Chen2019,mundhenk2021symbolic}, program synthesis \cite{Forstenlechner2018}, and combinatorial optimization problems \cite{Correa2022,Zhang2021book}.

\begin{figure}[!t]
  \centering
  \includegraphics[scale=0.4, viewport=10 10 550 240, clip=true]{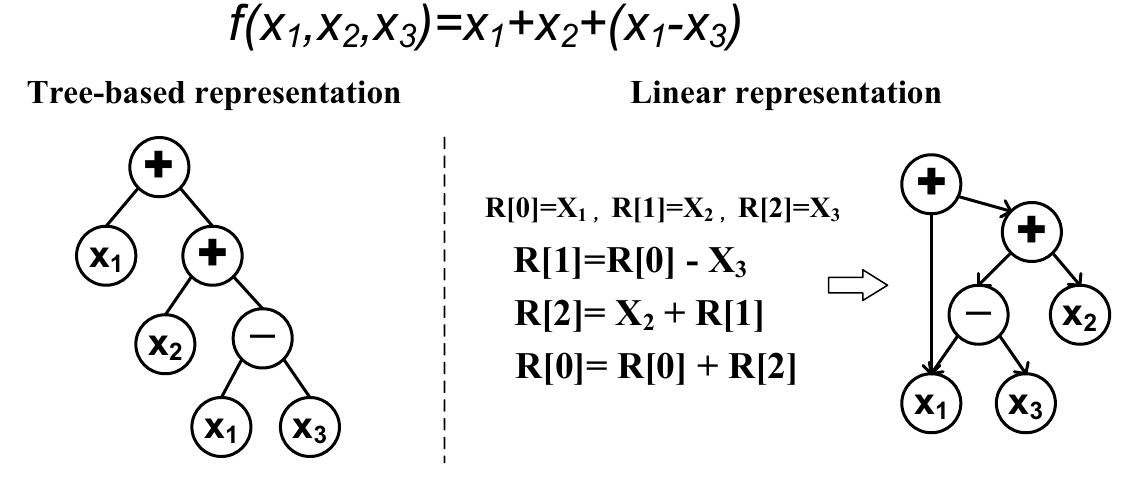}
  \caption{An example of GP individuals with tree-based and linear representations for the same mathematical formula.}\label{fig:treevslinear}
\end{figure}

LGP represents a computer program by a sequence of register-based instructions \cite{Brameier2007}.
In LGP, every instruction $f$ in the instruction sequence $\mathbf{F}=\{f_1,f_2,...f_{|F|}\}$ manipulates registers from the same set of registers $\mathbb{R}=\{\textrm{R}[0],\textrm{R}[1],\cdots,\textrm{R}[|\mathbb{R}|-1]\}$, based on the operation in the instruction (denoted as $f_{fun}$). The registers in $f$ can be categorized into destination registers ($f_{d}$) and source registers ($f_{s}$). In our work, there are at most one destination register and two source registers (denoted as $f_{s,1}$ and $f_{s,2}$) in each instruction. The final outputs of LGP programs are stored in designated destination registers, normally starting from the first register, $\textrm{R}[0]$, by default. An LGP program can be decoded into a graph. By connecting the operations in the instructions based on registers, the instruction sequence can be decoded into a directed acyclic graph (DAG), in which every graph node can have more than one parent. A directed edge points from a certain graph node to another providing inputs.
A comparison between a linear representation and a tree-based representation for the same mathematical formula ``$f(x_1,x_2,x_3)=x_1+x_2+(x_1 - x_3)$'' is shown in Fig. \ref{fig:treevslinear}.
Specifically, in the linear representation, $x_1$ to $x_3$ are read-only input registers, and the calculation registers $R[0]$ to $R[2]$ are initialized by $x_1$ to $x_3$ respectively (e.g., the first instruction is equivalent to ``$R[1]=x_1 - x_3$'' ). The final output of the instruction sequence is stored in $R[0]$. The DAG of the LGP individual is also shown in Fig. \ref{fig:treevslinear}.

Existing literature has shown that different GP representations have superior performance in different domains \cite{Sotto2021Graph,Kantschik2001,Wilson2008}. 
However, to the best of our knowledge, all of the GP methods evolve GP individuals only with one unified representation which essentially defines the search space and the corresponding fitness landscape. There is no existing literature that uses the search spaces from different representations to enhance the GP search for an optimization problem.
Since it is not guaranteed that the chosen representation must be effective for the specific problem, evolving multiple representations simultaneously reduces the risks of inadequate GP representation and is expected to be beneficial to improve GP performance.

\subsection{Enhancing Evolution By Switching Fitness Landscapes}
A fitness landscape consists of three components: search space, fitness function, and neighborhood function \cite{Pitzer2012}. The search space contains all possible solutions, the fitness function evaluates how good the solutions are, and the neighborhood function defines the neighborhood set of each solution.
This paper simultaneously evolves the GP individuals with multiple different representations, which is similar to simultaneously searching on multiple related fitness landscapes (as GP individual representations are highly related to fitness landscapes).
In the existing literature, there have been studies \cite{Wei2021,Wu2022} showing that making full use of related fitness landscapes in designing search mechanisms is an effective way to improve search performance.

Multitask optimization \cite{Ong2015} is an example of an optimization method that enhances search performance by mutually exchanging information among the fitness landscapes with similar fitness functions. Specifically, multitask optimization constructs similar fitness functions by simultaneously solving several similar tasks. For example, Gupta et al. \cite{Gupta2016}
used a multitask evolutionary computation method (i.e., multifactorial evolutionary algorithm) to simultaneously solve multiple continuous and discrete optimization problems whose optima are close in the search space. Yi et al. \cite{Yi2020} further confirmed that searching on similar tasks is also effective in solving combinatorial optimization problems.

Sharing the GP search information among tasks is helpful to GP evolution. For example, 
Zhong et al. \cite{Zhong2020MT-GEP} proposed a multifactorial gene expression programming method to solve more than one symbolic regression problem simultaneously.
Zhang et al. \cite{Zhang2021} compared the effectiveness of different evolutionary multitask frameworks in solving dynamic flexible job shop scheduling and extended the multitask frameworks to multi-objective optimization in flexible dynamic job shop scheduling \cite{Zhang2022MTMO}. Huang et al. \cite{HuangMTLGPinves} investigated the effectiveness of LGP in existing evolutionary multitask frameworks for solving dynamic job shop scheduling.

Simultaneously optimizing several alternative formulations for a single problem is another way to build up similar fitness landscapes, so-called multiform optimization.
For example, Da et al. \cite{Da2016} formulated a traveling salesman problem into a single-objective and a multi-objective optimization problem respectively, and solved the two optimization problems via a multitask optimization method. Since multi-objective formulations often introduce plateaus into fitness landscapes, it is expected that the multi-objective optimization task can remove some local optima from the single-objective formulation. Additional formulations can also be constructed by adding or relaxing constraints on the optimization problem \cite{Jiao2022}, which is equivalent to constructing different search spaces (and hence different fitness landscapes) for solving the single task.

Changing neighborhood functions to reshape the fitness landscape can construct related fitness landscapes in the search for effective solutions. One representative example is variable neighborhood search \cite{1997Hansen}, which switches neighborhood functions in the course of search. By searching within different neighborhoods, variable neighborhood search can reach distant solutions via local search and has a better chance to jump out from a local optimum. Variable neighborhood search is an effective strategy to enhance other search techniques \cite{Cazzaro2022}.

All of these existing studies show that designing search mechanisms based on fitness landscape considerations is beneficial to search performance.
However, using different solution representations to construct related fitness landscapes is not well investigated when solving the same task, especially in the GP area.
Although some studies of evolutionary multitask optimization use different solution representations for different tasks \cite{Feng2019,Feng2021}, their solution representations are mainly designed based on problem-specific decision variables, which is much more intuitive than designing GP representations.
Further, the solutions in most evolutionary computation studies are numerical and have a simple neighborhood function. This is vastly different from GP representations that are symbolic and might have neighborhoods exponentially increasing with program size. 

\subsection{Benefits of Switching GP Representations}
\begin{figure}
    \centering
    \includegraphics[scale=0.5, viewport=20 20 570 250, clip=true]{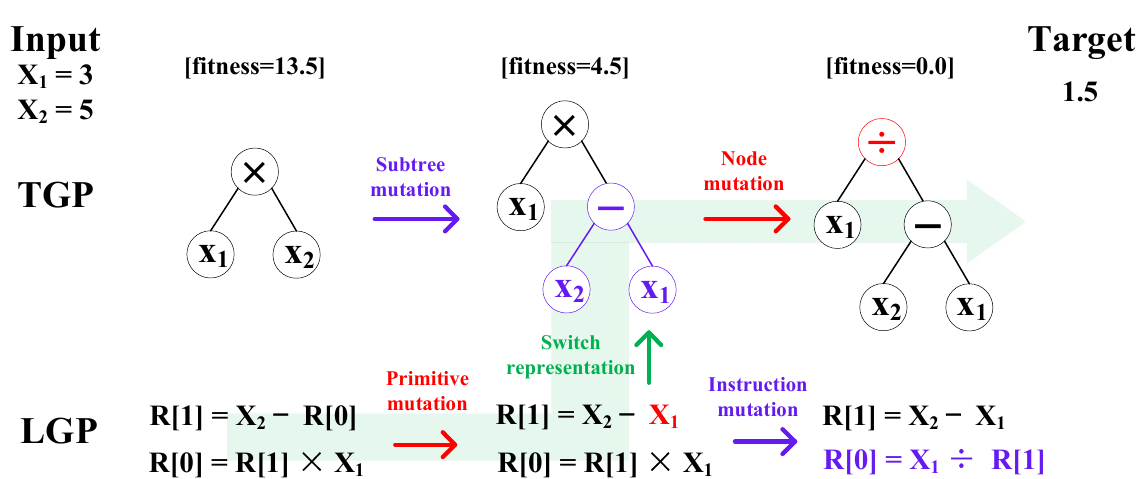}
    \caption{Switching GP representations to jump out of local optima.}
    \label{fig:switch}
\end{figure}

Utilizing the interplay among different GP representations is beneficial for GP evolution.
Fig. \ref{fig:switch} is an example to show the benefit of switching GP representations.
Given the inputs $X_1=3, X_2=5$ and the target output 1.5, we apply TGP and LGP to synthesize a mathematical formula. Each pair of TGP and LGP programs with the same fitness is representing the same program. Suppose TGP and LGP can only apply mutation operators to produce offspring and start from the same formula $f(X)=X_1 \times X_2$ whose fitness (defined as absolute error) is 13.5 ($|3\times5-1.5|=13.5$). The initial values of LGP registers in  Fig. \ref{fig:switch} are 0. 
To move on to the second step, LGP only needs to mutate the second primitive in the first instruction, but TGP has to perform a subtree mutation. Subtree mutation is a relatively large variation that leads to a large neighborhood. It is more likely (i.e., easier) for LGP to sample the offspring from a small neighborhood (i.e., by one-primitive mutation) than for TGP to sample the exact offspring from the large neighborhood (i.e., by subtree mutation). However, from the second to the third step, TGP only needs to mutate a tree node, while LGP has to mutate a new instruction. 
Fig. \ref{fig:switch} shows that switching solutions between LGP and TGP representations can share the search information in TGP and LGP. It is useful for GP individuals to reach better fitness via fewer and smaller variations and avoid difficult variations.

\section{Multi-representation Genetic Programming with TGP and LGP}
\subsection{Overall Framework}
\begin{figure}[!t]
 \centering
 \includegraphics[scale=0.52, viewport= 10 10 585 230, clip=true]{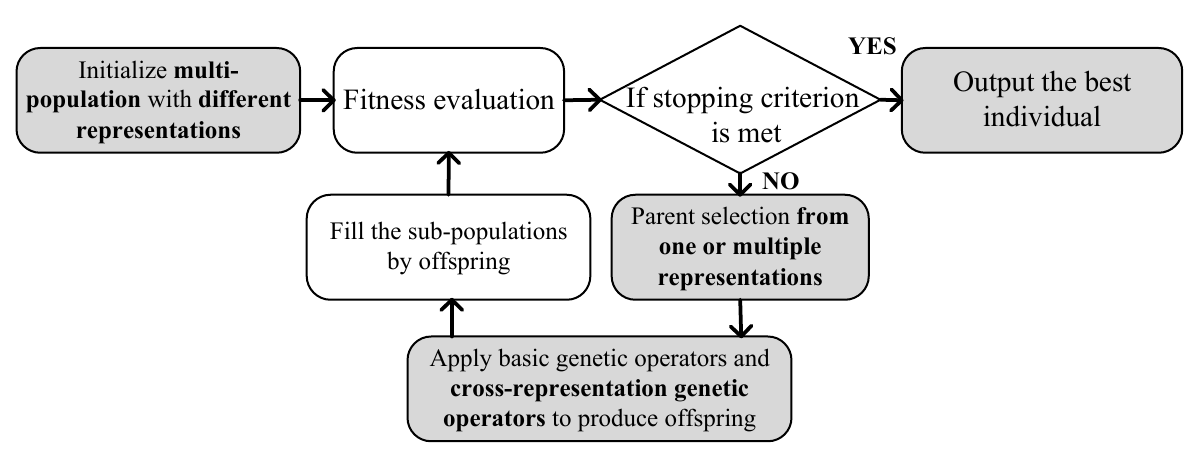}
 \caption{Evolutionary framework of MRGP. The novel components are highlighted by the dark boxes.}\label{fig:framework}
\end{figure}

%
%

\begin{algorithm}[!t]
\footnotesize
\caption{MRGP-TL}
\label{alg:framework}
\KwIn{cross-representation crossover rate $\theta_{t}$, tournament selection size $s$, maximum depth of the tree $\overline{d}$, maximum number of instructions $\overline{L}$, minimum number of instruction $\underline{L}$}
\KwOut{best individual $\mathbf{h}$}
Initialize two sub-populations, $\mathbb{S}_1$ for the tree-based representation and $\mathbb{S}_2$ for the linear representation.

\While{stopping criteria are not satisfied}{
    \tcp{Evaluation}
    Evaluate fitness of individuals $\forall\mathbf{f}\in\mathbb{S}_1 \bigcup \mathbb{S}_2$.

    Update the best individuals $\mathbf{h}$ in $\mathbb{S}_1 \bigcup \mathbb{S}_2$;

    \For{$j \leftarrow 1$ to $2$}{
    $\mathbb{S}_j'\leftarrow\emptyset$;

    Clone top-1\% individuals of $\mathbb{S}_j$ into $\mathbb{S}_j'$;

    \While{$|\mathbb{S}_j'| < |\mathbb{S}_j|$}{
        $rnd\leftarrow \texttt{rand}(0,1)$;

        \If{$rnd < \theta_{t}$}{
            $\mathbf{p}_1\leftarrow \texttt{TournamentSelection}(\mathbb{S}_j,s)$;

            $i\leftarrow \texttt{randint}(1,2)$;

            $\mathbf{p}_2\leftarrow \texttt{TournamentSelection}(\mathbb{S}_i,s)$;

            $\mathbf{c}\leftarrow \texttt{CALX}(\mathbf{p}_1, \mathbf{p}_2, \overline{d}, \overline{L}, \underline{L}) $;

        }
        \Else{
            Apply corresponding (i.e., TGP or LGP) basic genetic operators on $\mathbb{S}_j$ to produce offspring $\mathbf{c}$ (or $\mathbf{c}_1$ and $\mathbf{c}_2$);
        }

        $\mathbb{S}_j' \leftarrow \mathbb{S}_j' \bigcup \{\mathbf{c}\}$ (or $\mathbb{S}_j' \leftarrow \mathbb{S}_j' \bigcup \{\mathbf{c}_1, \mathbf{c}_2\}$);
    }
    $\mathbb{S}_j\leftarrow\mathbb{S}_j'$;
    }

%
%
%
%
%
%
%
%
%
}

\textbf{Return} $\mathbf{h}$.

\end{algorithm}

We propose an overall framework of the MRGP, as shown in Fig. \ref{fig:framework} (with the new components highlighted in grey). In contrast to the evolutionary framework of basic GP methods, MRGP evolves multiple sub-populations, each for a unique GP representation. When breeding offspring, MRGP selects parents from all the representations and also applies cross-representation genetic operators to produce offspring. Offspring of a certain representation fill the corresponding sub-population of the next generation. After generations of evolution, the best solution is output. Note that the best solution is either from the tree-based representation or the linear representation.

This paper studies MRGP based on the TGP and LGP, denoted as MRGP-TL.
The pseudo-code of MRGP-TL is shown in Alg. \ref{alg:framework} \footnote{$\texttt{rand}(a,b)$ returns a random floating-point number in $[a,b)$. $\texttt{randint}(a,b)$ returns a random integer number in $[a,b]$. $|\cdot|$ denotes the cardinality of a container (e.g., set or list). $(\cdot)$ following a container denotes getting an element from the container based on the index.}. First, MRGP-TL initializes two sub-populations, one for evolving tree-based GP individuals and the other for evolving LGP individuals. 
All individuals in these two sub-populations evolve simultaneously. 
For each sub-population, we perform elitism selection to retain elite individuals for the next generation (line 7).
To fill the sub-population of the next generation, 
we use tournament selection (i.e., $\texttt{TournamentSelection}(\cdot)$) to select individuals as parents and apply different genetic operators based on predefined rates. Specifically, MRGP-TL triggers the cross-representation adjacency-list based crossover ($\texttt{CALX}(\cdot)$) based on a predefined rate $\theta_{t}$ (line 10). If the cross-representation adjacency-list-based crossover is triggered, MRGP-TL selects a parent from the current sub-population and selects the other parent from one of the two sub-populations. $\texttt{CALX}(\cdot)$ accepts the two parents and produces an offspring. If the operator is not triggered, MRGP-TL applies basic TGP or LGP genetic operators to evolve tree-based and linear representations separately (lines 15-16).
The newly generated offspring form the new populations with different representations (line 17).
The evolution continues until a stopping criterion is met. The best individual among all the sub-populations with different representations is output as the final result.

\subsection{Cross-representation Adjacency List-based Crossover}
\begin{figure}[!t]
  \centering
  \includegraphics[scale=0.52, viewport=10 15 550 180, clip=true]{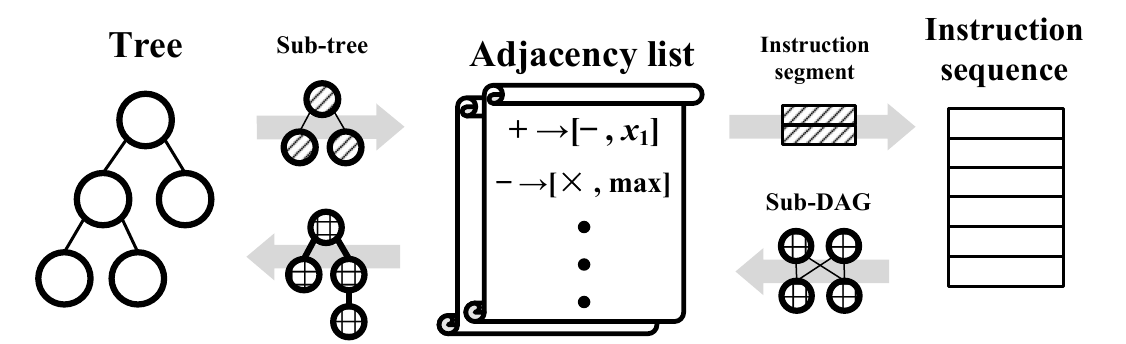}\\
  \caption{The schematic diagram of CALX between trees and instruction sequences}\label{fig:CALX}
\end{figure}

Knowledge transfer among representations is implemented by the cross-representation adjacency list-based crossover (CALX), as shown in Fig. \ref{fig:CALX}. To swap genetic materials, a tree or an instruction sequence first selects a sub-tree or an instruction segment. The instruction segment is essentially a sub-DAG (or multiple disconnected sub-DAGs).
The sub-tree and sub-DAGs are further converted into adjacency lists\footnote{Disconnected graph nodes are converted into adjacency lists with empty adjacent nodes $\textbf{A}$}. 
Based on the representation of the recipient, a new sub-tree or instruction segment is constructed based on the adjacency list and swapped into the recipient.

An adjacency list is a high-level representation of a graph. 
This paper denotes an adjacency list as
$$
\mathbf{L}=\left(\begin{array}{cccc}
\left[
fun_1,\mathbf{A}_1
\right]
&
\left[
fun_2,\mathbf{A}_2
\right]
&
\cdots
&
\left[
fun_{|\mathbf{L}|},\mathbf{A}_{|\mathbf{L}|}
\right]
\end{array}
\right)
$$
where each item $[fun_i,\mathbf{A}_i]$ specifies a function $fun_i$ and its adjacent nodes $\mathbf{A}_i$. Specifically, $\mathbf{A}_i$ contains one or two nodes in this paper since we only consider unary and binary functions.
For example, we convert the left tree in Fig. \ref{fig:treevslinear} as 
$$
\mathbf{L}=\left(\begin{array}{ccc}
\left[
+,\left[x_1, +\right]
\right]
&
\left[
+,\left[x_2, -\right]
\right]
&
\left[
-,\left[x_1, x_3\right]
\right]
\end{array}
\right)
$$
It is worth noting that the adjacency list in this paper uses primitive symbols (i.e., functions or terminals) to specify graph nodes to highlight building blocks, which is different from conventional adjacency lists which distinguish graph nodes by the indexes.


\begin{algorithm}[!t]
\footnotesize
\caption{$\texttt{CALX}$}
\label{alg:CALX}
\KwIn{Parent individuals $\mathbf{p}_1$ and $\mathbf{p}_2$, maximum depth of the tree $\overline{d}$, maximum number of instructions $\overline{L}$, minimum number of instruction $\underline{L}$}
\KwOut{An offspring $\mathbf{c}$}
Clone $\mathbf{p}_1$ as $\mathbf{c}$;

\If{$\mathbf{p}_1$ is a TGP individual}{\tcp{breeding trees based on adjacency lists}
    Randomly pick an inner tree node $t_1$ from $\mathbf{c}$;

    \If{$\mathbf{p}_2$ is a TGP individual}{
        Randomly pick an inner tree node $t_2$ from $\mathbf{p}_2$;

        $\mathbf{L} \leftarrow $ get the adjacency list of the sub-tree in $\mathbf{p}_2$ whose root is $t_2$;
    }
    \ElseIf{$\mathbf{p}_2$ is an LGP individual}{

        Randomly select a crossover point $t_2$ and select an instruction segment $\mathbf{F}' \subseteq [\mathbf{p}_2(t_2), \mathbf{p}_2(|\mathbf{p}_2|)]$;

        $\mathbf{L} \leftarrow $ get the adjacency list of the sub-graph from $\mathbf{F}'$;
    }

    $t'_1 \leftarrow \texttt{GrowTreeBasedAL}(\mathbf{L}, \text{the depth of }t_1\text{ in }\mathbf{c}, 1, \overline{d})$;

    Replace the sub-tree with the root of $t_1$ as the sub-tree with the root of $t'_1$ in $\mathbf{c}$;
}
\ElseIf{$\mathbf{p}_1$ is an LGP individual}{\tcp{breeding instructions based on adjacency lists}
    \If{$\mathbf{p}_2$ is a TGP individual}{
        Randomly select a crossover point $t_1$ and select an instruction segment $\mathbf{F}'_1 \subseteq [\mathbf{c}(1), \mathbf{c}(t_1)]$;

        $\mathbf{L}_1 \leftarrow $ get the adjacency list of the sub-graph from $\mathbf{F}_1'$;

        Randomly pick an inner tree node $t_2$ from $\mathbf{p}_2$;

        $\mathbf{L}_2 \leftarrow $ get the adjacency list of the sub-tree in $\mathbf{p}_2$ whose root is $t_2$;
    }
    \ElseIf{$\mathbf{p}_2$ is an LGP individual}{
        Randomly select a crossover point $t_1$ and select an instruction segment $\mathbf{F}'_1 \subseteq [\mathbf{c}(t_1), \mathbf{c}(|\mathbf{c}|)]$;

        $\mathbf{L}_1 \leftarrow $ get the adjacency list of the sub-graph from $\mathbf{F}'_1$;

        Randomly select a crossover point $t_2$ and select an instruction segment $\mathbf{F}'_2 \subseteq [\mathbf{p}_2(t_2), \mathbf{p}_2(|\mathbf{p}_2|)]$;

        $\mathbf{L}_2 \leftarrow $ get the adjacency list of the sub-graph from $\mathbf{F}'_2$;
    }

    %

    $\mathbf{c}\leftarrow\texttt{GrowInstructionBasedAL}(\mathbf{p}_1, \mathbf{L}_1, \mathbf{L}_2, t_1, n_1 )$

    \If{$|\mathbf{c}| \notin [\overline{L},\underline{L}]$}{
        $\mathbf{c}\leftarrow \mathbf{p}_1$;
    }
}

\textbf{Return} $\mathbf{c}$;
\end{algorithm}

\subsubsection{Breeding Trees Based on Adjacency Lists}
The pseudo-code of $\texttt{CALX}(\cdot)$ is shown in Alg.\ref{alg:CALX}. If the first and second parents are both TGP individuals, an inner tree node $t_2$ is randomly selected from the second parent, and an adjacency list $\mathbf{L}$ is generated based on the sub-tree under $t_2$ (lines 3-6). If the first parent is a TGP individual and the second parent is an LGP individual, we randomly select an instruction segment $\mathbf{F}'$ (lines 7-8) and convert it to sub-DAGs. $\mathbf{L}$ is further constructed based on the selected sub-graphs (line 9). Then, we apply $\texttt{GrowTreeBasedAL}(\cdot)$ to build a sub-tree based on $\mathbf{L}$, as shown in Alg. \ref{alg:growTree}.

$\texttt{GrowTreeBasedAL}(\cdot)$ is a recursive function to construct tree nodes based on $\mathbf{L}$. Specifically, if $\texttt{GrowTreeBasedAL}(\cdot)$ accepts an empty $\mathbf{L}$ or has reached the maximum depth, it returns a random sub-tree to ensure the validity (lines 1-2). Otherwise, $\texttt{GrowTreeBasedAL}(\cdot)$ grows a tree node $r$ based on $\mathbf{L}$ (line 3). If $r$ is a function,  $\texttt{GrowTreeBasedAL}(\cdot)$ checks the adjacency list and recursively applies $\texttt{GrowTreeBasedAL}(\cdot)$ to grow the sub-trees of $r$ (lines 4-13). Random sub-trees are constructed if there are no consistent entities in $\mathbf{L}$ (lines 11-12).

\begin{algorithm}[!t]
\footnotesize
\caption{$\texttt{GrowTreeBasedAL}$}
\label{alg:growTree}
\KwIn{Adjacency list $\mathbf{L}$, current depth $d$, index of $\mathbf{L}$ $I$, maximum depth of the tree $\overline{d}$}
\KwOut{A tree root $r$}
\If{$|\mathbf{L}|=0$ or $d=\overline{d}$}{
    \textbf{Return} $r \leftarrow$ a random sub-tree whose depth $\leq\overline{d}-d+1$;
}


    $[r, \mathbf{A}] \leftarrow \mathbf{L}(I)$;

    \If{$r$ is a function}{
        \For{$j \leftarrow 1 \text{ to } |\mathbf{A}|$}{
            $c'\leftarrow \mathbf{A}(j)$;

            \If{$c'$ is a function}{
                $\mathbf{L'}\leftarrow$ collect the entities from $\mathbf{L}(k), k\in [I,|\mathbf{L}|]$ with $\mathbf{L}(k).fun=c'$;


                \If{$\mathbf{L'}\neq\emptyset$}{
                    $c' \leftarrow \texttt{GrowTreeBasedAL}(\mathbf{L}, d+1, \texttt{randint}(1,|\mathbf{L'}|), \overline{d})$;
                }
                \Else{
                    $c' \leftarrow$ a random sub-tree whose depth $\leq\overline{d}-d-1$;
                }
            }

            Append $c'$ as $r$'s child;
        }

    }

\textbf{Return} $r$;
\end{algorithm}

\begin{figure}
    \centering
    \includegraphics[scale=0.45, viewport=10 15 550 170, clip=true]{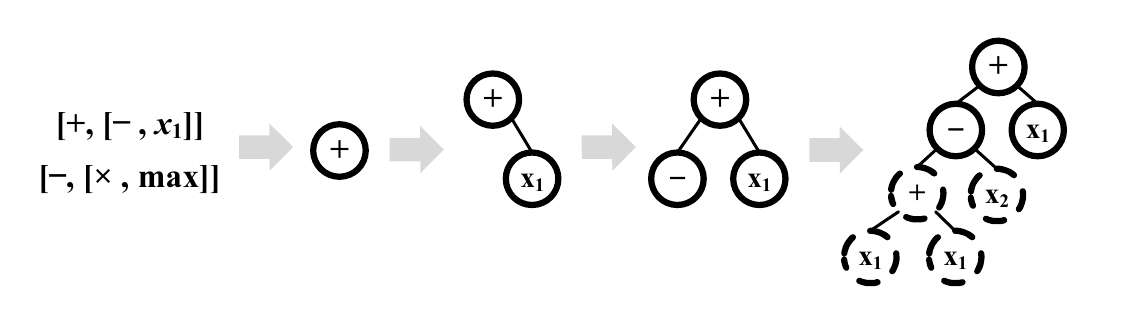}
    \caption{An example of constructing a tree by $\texttt{GrowTreeBasedAL}(\cdot)$. The dashed tree nodes are randomly generated.}
    \label{fig:constructTree}
\end{figure}

Fig. \ref{fig:constructTree} is an example of constructing a tree based on an adjacency list. The first item in the adjacency list is ``$[+,[-, x_1]]$'', and hence the root node of the new sub-tree is ``$+$''. Since the adjacent nodes of ``$+$'' are ``$-$'' (a function) and ``$x_1$'' (a terminal), we append ``$x_1$'' to the ``$+$'' and recursively apply $\texttt{GrowTreeBasedAL}(\cdot)$ with the second item (i.e., $[-,[\times,\max]]$) in the adjacency list to grow the sub-tree since the function of the second item ``$-$'' is coincident with the function adjacent node in the first item. Since the adjacent nodes of ``$-$'' (i.e., ``$\times$'' and ``$\max$'') are not included as items in the adjacency list, we randomly generate the sub-trees under ``$-$''.   

\subsubsection{Breeding Instructions Based on Adjacency Lists}
In $\texttt{CALX}(\cdot)$, if the first parent $\mathbf{p}_1$ is an LGP individual, sub-tree and sub-DAGs are respectively selected based on parents' representation, a sub-tree for TGP parent and sub-DAGs for LGP parent (lines 14, 16, 19, and 21). Specifically, the sub-DAGs are selected by selecting an instruction segment from the LGP parent.
Then, adjacency lists $\mathbf{L}_1$ and $\mathbf{L}_2$ are constructed respectively based on the selected sub-tree and sub-graphs (lines 15, 17, 20 and 22).
A new instruction sequence is constructed and swapped into $\mathbf{p}_1$ to produce offspring by $\texttt{GrowInstructionBasedAL}(\cdot)$ (line 23).

CALX applies $\texttt{GrowInstructionBasedAL}(\cdot)$ to construct a new instruction segment for LGP, as shown in Alg. \ref{alg:growInstruction}. First, $|\mathbf{L}_1|$ instructions are randomly removed (line 1).
Then an insertion point $s$ is selected for inserting the new instruction segment (line 2).
Instructions are sequentially constructed based on $\mathbf{L}_2$ and swapped into the program context (lines 3-6).
To connect the functions and maintain the topological structure of the functions based on $\mathbf{L}_2$, we check the instruction sequence reversely (i.e., from the top of the graph to the bottom) (lines 7-21) so that every newly generated instruction 1) is effective to the final output by altering the destination registers $\mathbf{c}(j)_{d}$ (lines 9-10), and 2) accepts the inputs from corresponding functions and constants based on $\mathbf{L}_2$ by altering the source registers $\mathbf{c}(j)_{s, g}$ (lines 11-21).
Specifically, the effectiveness of an instruction is checked by an $\mathcal{O}(n)$ algorithm \cite{Brameier2007} (line 9). If the selected instruction is not effective, we randomly mutate the destination register of the instruction until it is effective.
$\texttt{GrowInstructionBasedAL}(\cdot)$ assigns source registers based on the adjacent node $b$ (line 12). If $b$ is a function, we set the source register of the selected instruction $\mathbf{c}(j)$ as the destination register of a random instruction whose function is coincident with $b$ (lines 14-16). If there is no such an instruction, we set the source register as the destination register of a random instruction precedent to $\mathbf{c}(j)$ (lines 18-19). The constant adjacent nodes replace the source registers directly (lines 20-21).

\begin{algorithm}[!t]
\footnotesize
\caption{$\texttt{GrowInstructionBasedAL}$}
\label{alg:growInstruction}
\KwIn{An LGP individual $\mathbf{c}$, adjacency list of the first parent $\mathbf{L}_1$, adjacency list of the second parent $\mathbf{L}_2$, crossover point $t_1$, instruction range $n_1$}
\KwOut{The LGP offspring $\mathbf{c}$}
Randomly remove $|\mathbf{L}_1|$ instructions from $[\mathbf{c}(t_1),\mathbf{c}(t_1+n_1)]$;

$s\leftarrow t_1+\texttt{randint}(n_1 - |\mathbf{L}_1|)$;

\tcp{Construct an instruction list}
\For{$j\leftarrow 1$ to $|\mathbf{L}_2|$}{
    $[a, \mathbf{A}] \leftarrow \mathbf{L}_2(j)$;

    $f \leftarrow $ randomly generate an instruction whose function is $a$;

%
    \tcp{Swap into the program context}
    Insert $f$ to $\mathbf{c}(s)$; 
}

\tcp{Assign registers}
\For{$j \leftarrow s+|\mathbf{L}_2|-1$ to $s$}{
    $[a, \mathbf{A}] \leftarrow \mathbf{L}_2(s+\mathbf{L}_2-j)$;

    \tcp{Assign destination registers}
    \If{$\mathbf{c}(j)$ is not effective to the final output}{
        Randomly mutate $\mathbf{c}(j)_{d}$ until $\mathbf{c}(j)$ is effective;
    }

    \tcp{Assign source registers}
    \For{$g \leftarrow 1 \text{ to } |\mathbf{A}|$}{
        $b\leftarrow \mathbf{A}(g)$;

        \If{$b$ is a function}{
%

            $\mathbf{L'}\leftarrow$ collect the entity indices from $[j,s]$ where $\mathbf{L}_2(k).fun=b$ and $k\in[j,s]$;

            \If{$\mathbf{L'}\neq\emptyset$}{

                $\mathbf{c}(j)_{s, g} \leftarrow \mathbf{c}(\mathbf{L'}(\texttt{randint}(1,|\mathbf{L'}|)))_{d}$;
            }
            \Else{
                \If{$j>0$ and $\texttt{randint}(0,j)-1>0$}{

                    $\mathbf{c}(j)_{s, g} \leftarrow \mathbf{c}(\texttt{randint}(1, j))_{d}$;
                }
            }
        }
        \ElseIf{$b$ is a constant}{
            $\mathbf{c}(j)_{s, g} \leftarrow b$;
        }
    }
}

\textbf{Return} $\mathbf{c}$;
\end{algorithm}

\begin{figure}
    \centering
    \includegraphics[scale=0.47, viewport=10 190 550 320, clip=true]{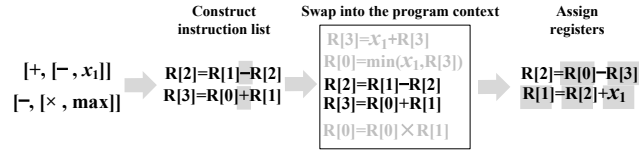}
    \caption{An example of constructing instructions by $\texttt{GrowInstructionBasedAL}(\cdot)$. Shadowed primitives are the focus of each step.}
    \label{fig:constructInstruction}
\end{figure}

Fig. \ref{fig:constructInstruction} shows an example of constructing an instruction list based on the adjacency list. First, we construct an instruction list in which the functions (i.e., ``$+$'' and ``$-$'') are specified by the adjacency list. Note that the order of functions in the instruction list is reversed to the order in the adjacency list since LGP programs output final results from the bottom.
Second, we swap the newly constructed instruction list into the program context.
Third, we adjust the registers in the newly constructed instruction list to maintain the adjacency relationship in the offspring. In this example, we change the destination register R[3] into R[1] to ensure the new instruction list to be effective in the offspring, change R[0] in the second instruction into R[2] and change R[1] into $x_1$ to fulfill the adjacency relationship ``$+ \rightarrow[-,x_1]$''. To connect the newly constructed instruction list with the precedent instructions in existing programs, the source registers in the first instruction are also updated.

\section{Empirical Studies of MRGP-TL}
We verify the effectiveness of MRGP-TL by two applications, symbolic regression and automatic design of decision rules in dynamic combinatorial optimization problems. 
Symbolic regression is a supervised learning problem, in which GP learns regression models to map the input features to given target outputs without presuming the model structure. GP has shown great success in solving symbolic regression problems \cite{Mundhenk2021,Piringer2022,Zhang2023}.
Automatically designing decision rules for dynamic combinatorial optimization problems uses GP to automatically learn decision rules to make instant reactions for dynamic events in combinatorial optimization problems. 
Unlike symbolic regression problems where there are target outputs for training, the decision rule design problems have no target outputs available. GP methods have to search for solutions based on a black-box performance indicator (e.g., the performance of simulations).
Specifically, we focus on dynamic job shop scheduling (DJSS) as an example of dynamic combinatorial optimization in this paper.
DJSS is a common and important combinatorial optimization problem in real-world practice.
Designing instant decision rules for DJSS is a challenging problem for all GP representations \cite{HuangMTLGPinves,FAN2021105401}.

\subsection{Comparison Design}\label{sec:compdesg}
To verify the effectiveness of MRGP-TL, we compare MRGP-TL with three baseline methods in these two applications. The first two are the basic TGP and LGP. Then, a baseline GP method with two independent sub-populations is developed (denoted as ``TLGP''). The two sub-populations independently evolve tree-based and linear representations by the basic genetic operators for each representation and do not exchange genetic materials among representations. 
To ensure fairness, we set the parameters of the compared GP methods for the two applications respectively, following the popular settings in existing literature \cite{Koza1992,Chen2019instance,Huang2022}. The parameter settings for the two applications are demonstrated in Sections \ref{sec:SRparams} and \ref{sec:DJSSparams}.

\subsection{Application I: Symbolic Regression}\label{sec:SR}
\subsubsection{Problem Description}
In this section, we apply MRGP-TL to symbolic regression problems. 
We select three synthetic benchmarks and five real-world benchmarks, as shown in Table \ref{tb:benchmark2}. The benchmarks are selected from recently published papers for solving symbolic regression \cite{Huang2022SLGP,Al-Helali2021}. The ground truth functions of the synthetic benchmarks cover a wide range of functions (e.g., $\times$ and $\sin$), and the real-world benchmarks have various numbers of features and data ranges. 

This paper applies relative square error (RSE) to measure the performance of GP methods, as shown in Eq. \ref{eq:rse}.
\begin{equation}
\label{eq:rse}
    \textrm{RSE}= \frac{\text{MSE}(\textbf{y},\textbf{\^{y}})}{\textrm{VAR}(\textbf{y})}=\frac{\sum_i^n (y_i - \hat{y_i})^2}{\sum_i^n (y_i - \overline{y})^2}
\end{equation}
where MSE is the mean square error, VAR is the variance, and $\textbf{y}$ and $\textbf{\^{y}}$ are the target output and estimated output respectively. $\overline{y}$ is the average of the target output. A small RSE value implies that a regression model has a good fitting performance with the given data.

\begin{table}
\centering
\caption{The symbolic regression problems}
\label{tb:benchmark2}
\scalebox{0.95}{
\setlength{\tabcolsep}{5pt}
\renewcommand{\arraystretch}{0.2}
\scriptsize
\begin{tabular}{@{}m{13mm} | m{30mm}| m{9mm}| m{9mm}| m{13mm}}\hline
\makecell[c]{Benchmarks} & \makecell[c]{Function} & \makecell[c]{\#Features} & \makecell[c]{Data\\ range} & \makecell[c]{\#Points\\(Train,Test)} \\\hline
\multicolumn{5}{c}{Synthetic benchmarks}\\\hline
\makecell[c]{Nguyen4} & $f(x)=x^6+x^5+x^4+x^3+x^2+x$ & \makecell[c]{1} & \makecell[c]{[-1,1]} & \makecell[c]{(20,1000)}\\


\makecell[c]{Keijzer11} & $f(x,y)=xy+\sin((x-1)(y-1))$ & \makecell[c]{2} & \makecell[c]{[-1,1]} & \makecell[c]{(100,900)}\\

\makecell[c]{R1} & $f(x)=\frac{(x+1)^3}{x^2-x+1}$ & \makecell[c]{1} & \makecell[c]{[-2,2]} & \makecell[c]{(20,1000)}\\ \hline

\multicolumn{5}{c}{Real-world benchmarks}\\\hline

\makecell[c]{Airfoil} & unknown & \makecell[c]{5} & \makecell[c]{-} & \makecell[c]{(1127,376)}\\

\makecell[c]{BHouse} & unknown & \makecell[c]{13} & \makecell[c]{-} & \makecell[c]{(380,126)}\\

\makecell[c]{Tower} & unknown & \makecell[c]{25} & \makecell[c]{-} & \makecell[c]{(3749,1250)}\\

\makecell[c]{Concrete} & unknown & \makecell[c]{8} & \makecell[c]{-} & \makecell[c]{(772,258)}\\ 

\makecell[c]{Redwine} & unknown & \makecell[c]{11} & \makecell[c]{-} & \makecell[c]{(1199, 400)} \\
\hline

\end{tabular}
}
\end{table}

\subsubsection{Parameter Settings}\label{sec:SRparams}
In symbolic regression problems, LGP evolves a population with 256 individuals for 200 generations. LGP applies linear crossover, effective macro mutation, effective micro mutation, and reproduction in breeding \cite{Brameier2007}, with a genetic operator rate of 30\%:30\%:30\%:10\% respectively. Each LGP individual has at most 100 instructions and manipulates 8 registers. 
TGP evolves a population with 1024 individuals for 50 generations, and applies crossover, mutation, and reproduction in breeding, with a genetic operator rate of 80\%:15\%:5\% respectively. Each TGP individual has a maximum tree depth of 10.

For the two algorithms with multiple representations (i.e., TLGP and MRGP-TL), each sub-population has 128 individuals and evolves 200 generations. 
The parameters of the MRGP-TL are defined based on the baseline method. Specifically, the knowledge transfer rate is defined as 30\% by default, without out loss of generality. Since the proposed adjacency list-based operators which are used to transfer knowledge among sub-populations can also exchange the genetic materials for the same representation, the LGP sub-population in MRGP-TL does not apply linear crossover operator, and the TGP sub-population in MRGP-TL reduces the crossover rate from 80\% to 50\%.
All the compared methods apply a tournament selection with a size of 7 to perform parent selection and apply an elitism selection with an elitism rate of 10\% to retain elite individuals. 
The other parameters of TLGP and MRGP-TL are kept the same as in the basic TGP and LGP methods.

All the compared GP methods use the same function set and terminal set (LGP methods have registers in the terminal set additionally). The function set includes 8 functions, which are \{$+,-,\times,\div,\sin,\cos,\ln(|\cdot|),\sqrt{|\cdot|}$\}\footnote{$\div$ returns 1.0 if the dividend equals to 0.0. $\ln(|\cdot|)$ returns the operand if the raw output is smaller than $-50$.}. The input feature set is defined based on the inputs of benchmark problems. 


\subsection{Application II: Dynamic Job Shop Scheduling Problems}\label{sec:DJSS}



\subsubsection{Problem Description}
This section applies MRGP-TL to design decision rules (i.e., dispatching rules) for DJSS \cite{Nguyen2017a,Zhang2021book}. 
We focus on the DJSS problems with new job arrival, in which jobs come into the job shop over time. 
A DJSS problem has a set of jobs $\mathcal{J}$ and a set of machines $\mathcal{M}$. Each $j\in\mathcal{J}$ consists of a sequence of operations $\mathbf{O}_j=\{o_{j1},o_{j2},...,o_{jl_j}\}$ where $l_j$ is the number of operations in job $j$. Every $o_{ji}$ can only be processed after $o_{j,i-1}$ is finished $(2\leq i\leq l_j)$. Each job $j$ has an arrival time $\alpha_j$, a due date $d_j$, and a weight $\omega_j$. $o_{ji}(1\leq i \leq l_j)$ is going to be processed by a specific machine $\pi(o_{ji})\in\mathcal{M}$ with a positive processing time $\delta(o_{ji})$. Every machine can only process one operation at any time, and the execution of the operation cannot be interrupted by other operations. 

A DJSS simulation is built up based on the description, and the settings of the DJSS simulations are designed based on the existing literature \cite{Huang2021LGP4DJSS}. 
Specifically, there are 10 machines in the job shop. Each job has 2 to 10 operations, each operation with a processing time ranging from 1 to 99.
To evaluate the performance of GP individuals in a steady job shop, we warm up the job shop with the first 1000 jobs and only take the subsequent 5000 jobs into account when evaluating GP individuals.

Jobs come into the job shop based on a Poisson distribution, as shown in Eq. \ref{eq:poisson}. $t$ is the time interval before next job arrival. $\lambda$ is the mean processing time of a job in the job shop, defined by Eq. \ref{eq:lambda}. $\nu$ is the average number of operations in the jobs, and $\mu$ is the average processing time of operations. 
The utilization level of machines $\rho$ defines the arrival rate of jobs. A large $\rho$ implies that jobs will be processed by the job shop very quickly (i.e., a small mean actual processing time of jobs) and that new jobs arrive to the job shop in a shorter time.
\begin{equation}\label{eq:poisson}
  P(t=\textrm{next job arrival time})\sim\exp(-\frac{t}{\lambda})\\
\end{equation}
\begin{equation}\label{eq:lambda}
  \lambda = \frac{\nu\cdot \mu}{\rho \cdot |\mathbf{\mathcal{M}}|}
\end{equation}

The simulation performance is seen as the performance of GP individuals.
There are six optimization objectives for the job shop in our work, which are formulated as follows. $T_{max}$ and $F_{max}$ denote the maximum tardiness and flowtime among all the jobs respectively. $c_j$, $d_j$, and $a_j$ denote the completion time, the due date, and the arrival time of job $j$. 
$T_{mean}$ and $F_{mean}$ denote the mean tardiness and flowtime over all the jobs respectively. $WT_{mean}$ and $WF_{mean}$ denote the weighted mean tardiness and flowtime respectively. The maximum and mean objectives comprehensively measure the worst and average performance of all the compared methods.
\begin{enumerate}
  \item $T_{max} = \max_{j\in\mathcal{J}}(\max(c_j - d_j, 0))$
  \item $T_{mean} = \frac{\sum_{j\in\mathcal{J}}(\max(c_j-d_j, 0))}{|\mathcal{J}|}$
  \item $WT_{mean} = \frac{\sum_{j\in\mathcal{J}}(\omega_j \times \max(c_j-d_j, 0))}{|\mathcal{J}|}$
  \item $F_{max} = \max_{j\in\mathcal{J}}(c_j - a_j)$ 
  \item $F_{mean} = \frac{\sum_{j\in\mathcal{J}}(c_j - a_j)}{|\mathcal{J}|}$
  \item $WF_{mean} = \frac{\sum_{j\in\mathcal{J}}\omega_j(c_j - a_j)}{|\mathcal{J}|}$
\end{enumerate}
To comprehensively verify the performance of the proposed method on different difficulty levels, two utilization levels (i.e., 0.85 and 0.95) are adopted for the simulation. A higher utilization level implies a busier job shop and more difficulty to find a schedule. In short, this paper tests twelve scenarios which are notated by ``$\langle$Objective, Utilization level$\rangle$''. The twelve scenarios are $\langle T_{max},0.85 \rangle$, $\langle T_{max},0.95 \rangle$, $\langle T_{mean},0.85 \rangle$, $\langle T_{mean},0.95 \rangle$, $\langle WT_{mean},0.85 \rangle$, $\langle WT_{mean},0.95 \rangle$, $\langle F_{max},0.85 \rangle$, $\langle F_{max},0.95 \rangle$, $\langle F_{mean},0.85 \rangle$, $\langle F_{mean},0.95 \rangle$, $\langle WF_{mean},0.85 \rangle$, and $\langle WF_{mean},0.95 \rangle$.

\subsubsection{Parameter Settings}\label{sec:DJSSparams}
The parameters of the compared methods are set based on the common settings in using GP methods to design dispatching rules for DJSS problems \cite{Zhang2021book,Huang2022}.
Specifically, all the TGP individuals have a maximal tree depth of eight, and all the LGP individuals have at least one instruction and at most fifty instructions.
The rest of the parameters are kept the same as in Section \ref{sec:SRparams}.
In DJSS problems, all the compared methods adopt the function set (i.e., $\{+,-,\times,\div,\max,\min\}$) and the terminal set in Table \ref{tb:terminals}.


\begin{table}[!t]
 \centering
 \caption{The terminal set}\label{tb:terminals}
 \scalebox{0.85}{
 \begin{tabular}{p{13mm}|p{60mm}|p{13mm}|p{48mm}}\hline
 Notation & Description                                                                                                                               & Notation & Description                                                                           \\ \hline
PT       & Processing time of an operation in a job                                                                                                  & W        & Weight of a job                                                                       \\
NPT      & Processing time of the next operation for a certain operation in a job                                                                    & rDD      & Difference between the given due   date of a job and the system time                  \\
WINQ     & Total processing time of the operations in a   machine buffer. The machine is the corresponding machine for the next   operation in a job & NWT      & Waiting time of the next   to-be-ready machine                                        \\
WKR      & Total remaining processing time of a   job                                                                                                & TIS      & Difference between system time and   the job arrival time                        \\
rFDD     & Difference between the given due date of an   operation and the system time                                                               & SL       & Slack: difference between the   given due date and the sum of the system time and WKR \\
OWT      & Waiting time of an operation                                                                                                              & NIQ      & Number of operations in a machine   buffer                                            \\
NOR      & Number of remaining operations of a job                                                                                                   & WIQ      & Total processing time of   operations in a machine buffer                             \\
NINQ     & Number of operations in the buffer of a   machine which is the corresponding machine of the next operation in a job                       & MWT      & Waiting time of a machine                                                           \\ \hline
 \end{tabular}
 }
\end{table}

For each independent run, a GP method first evolves on the training set and produces a final output rule based on a validation set with 10 DJSS instances. The rule with the best performance on the validation set is tested on 50 unseen DJSS instances. The test performance is defined as the mean performance on these test instances. The GP methods are trained on one DJSS instance for each generation, and the DJSS training instances are rotated every generation to improve the generalization ability of GP rules \cite{Hildebrandt2010}. All the compared methods have the same maximum number of simulations (i.e., fitness evaluation) in training.

\subsection{Empirical Results}
This section analyzes the test and training performance of the compared methods for solving the symbolic regression and DJSS problems. 
\subsubsection{Test Performance}
Table \ref{tb:testperformance} shows the average test performance of the compared methods in solving the two kinds of problems. We perform a Friedman test ($\alpha=0.05$) with a Bonferroni correction on the test performance of the compared methods. The null hypothesis of the Friedman test is that there is no significant difference in the test performance of the compared methods. 

The p-value of the Friedman test is 0.0015, which indicates a significant difference (i.e., alternative hypothesis) among the compared methods. Moreover, MRGP-TL has the best (i.e., smallest) mean rank of test performance among all the compared methods, with very promissing pair-wise comparison p-values with other compared methods. The results and statistical analyses confirm that the proposed MRGP-TL has a significantly better overall performance than the other three compared methods.

To further investigate the effectiveness of the compared methods on different datasets, Table \ref{tb:testperformance} shows the results of Wilcoxon rank-sum test with Bonferroni correction and an $\alpha$ of 0.05 over the test performance of the compared methods. $+$, $-$, and $\approx$ denote that a certain compared method is significantly better than, worse than, or performs similarly to the proposed MRGP-TL respectively, based on the Wilcoxon rank-sum test. The best mean performance is highlighted in bold font. We see that in most datasets and scenarios, MRGP-TL has a very competitive performance with the compared methods. More specifically, MRGP-TL has the best mean performance on 12 of 20 datasets and scenarios.
The results confirm that sharing knowledge between tree-based and linear representation successfully improves the effectiveness of GP methods.

\begin{table}
\caption{The mean test performance (std.) of the compared methods}
\label{tb:testperformance}
\footnotesize
\centering
\scalebox{0.8}{
\setlength{\tabcolsep}{3pt}
\begin{tabular}{c|c|c|c||c} \hline
\makecell[c]{Datasets or\\scenarios}       & TLGP          & TGP           & LGP           & MRGP-TL      \\\hline
\multicolumn{5}{c}{Test RSE (std.) of Symbolic Regression} \\\hline
Nguyen4       & 0.069 (0.059) $-$           & 0.053 (0.091) $\approx$           & 0.149 (0.248) $-$           & \textbf{0.051} (0.087)               \\
Keijzer11     & 0.365 (0.240) $\approx$           & \textbf{0.273} (0.121) $\approx$           & 0.339 (0.142) $\approx$           & 0.323 (0.133)                \\
R1            & 0.035 (0.029) $\approx$           & \textbf{0.022} (0.023) $\approx$           & 0.034 (0.035) $\approx$           & 0.025 (0.025)               \\
Airfoil       & 0.667 (0.091) $\approx$           & \textbf{0.638} (0.117) $\approx$           & 0.643 (0.132) $\approx$           & 0.643 (0.098)               \\
Bhouse        & 0.384 (0.100) $-$             & 0.392 (0.131) $\approx$           & 0.404 (0.126) $-$           & \textbf{0.325} (0.076)               \\
Tower         & 0.358 (0.052) $-$           & 0.364 (0.053) $-$           & 0.345 (0.046) $-$           & \textbf{0.325} (0.037)               \\
Concrete      & 0.496 (0.096) $-$           & 0.438 (0.107) $\approx$           & 0.471 (0.099) $-$           & \textbf{0.39} (0.078)                \\
Redwine       & 0.745 (0.042) $\approx$           & 0.761 (0.036) $\approx$           & 0.759 (0.034) $\approx$           & \textbf{0.757} (0.035)               \\\hline
\multicolumn{5}{c}{Test Objective Values (std.) of DJSS} \\\hline
$\langle Tmax,0.85\rangle$                               & 1939.8 (50.4) $\approx$          & 1928.4 (40.4) $\approx$        & 1956.3 (53.8) $-$       & \textbf{1926.9 (79.3)} \\
$\langle Tmax,0.95\rangle$                               & 4009.2 (98.9) $-$          & 4060.6 (116) $-$         & 3999.2 (90.9) $\approx$       & \textbf{3964.2 (89)}   \\
$\langle Tmean,0.85\rangle$                              & \textbf{417.0 (3.2) $\approx$}     & 417.3 (2.5) $\approx$          & 417.9 (2.3) $\approx$         & 417.2 (3.4)            \\
$\langle Tmean,0.95\rangle$                              & 1116.3 (9.3) $\approx$           & \textbf{1116.2 (10) $\approx$} & 1118.2 (10.7) $\approx$       & 1116.5 (10.6)          \\
$\langle WTmean,0.85\rangle$                             & 725.8 (6.1) $\approx$            & 727.5 (6.5) $-$          & 724.3 (5.4) $\approx$         & \textbf{724.0 (5.7)}     \\
$\langle WTmean,0.95\rangle$                             & 1730.4 (23.6) $\approx$          & 1747.4 (29.6) $-$        & 1729.6 (27.7) $\approx$       & \textbf{1723.5 (19.4)} \\
$\langle Fmax,0.85\rangle$                               & 2506.6 (50.3) $\approx$          & 2494.3 (30) $\approx$          & 2509.8 (58.8) $\approx$       & \textbf{2493.0 (33.3)}   \\
$\langle Fmax,0.95\rangle$                               & \textbf{4544.3 (98.5) $\approx$} & 4572.3 (96.5) $\approx$        & 4585.4 (126.1) $\approx$      & 4553.0 (110.8)           \\
$\langle Fmean,0.85\rangle$                              & 864.0 (3.2) $\approx$              & 863.2 (4.2) $\approx$          & 864.7 (4) $\approx$           & \textbf{862.8 (2.9)}   \\
$\langle Fmean,0.95\rangle$                              & \textbf{1564.9 (10.3) $\approx$} & 1565.4 (8.5) $\approx$         & 1566.8 (10.8) $\approx$       & 1565.3 (10.6)          \\
$\langle WFmean,0.85\rangle$                             & 1704.0 (10.2) $\approx$            & 1705.4 (7.5) $\approx$         & \textbf{1702.6 (7) $\approx$} & 1702.8 (5.6)           \\
$\langle WFmean,0.95\rangle$                             & 2718.4 (26.4) $\approx$          & 2730.1 (29.3) $-$        & 2715.8 (16.4) $\approx$       & \textbf{2711.0 (20.5)}   \\\hline
win-draw-lose & 0-15-5         & 0-15-5         & 0-15-5         &        \\\hline      
Mean rank & 2.7     & 2.7   & 3.05 & \textbf{1.55}  \\\hline
\makecell[c]{p-value\\(vs. MRGP-TL)}   & 0.029 & 0.029   & 0.001  &        \\\hline
\end{tabular}}
\end{table}

\subsubsection{Training Performance}
\begin{figure}
    \centering
    \includegraphics[scale=0.72, viewport=15 30 400 280, clip=true]{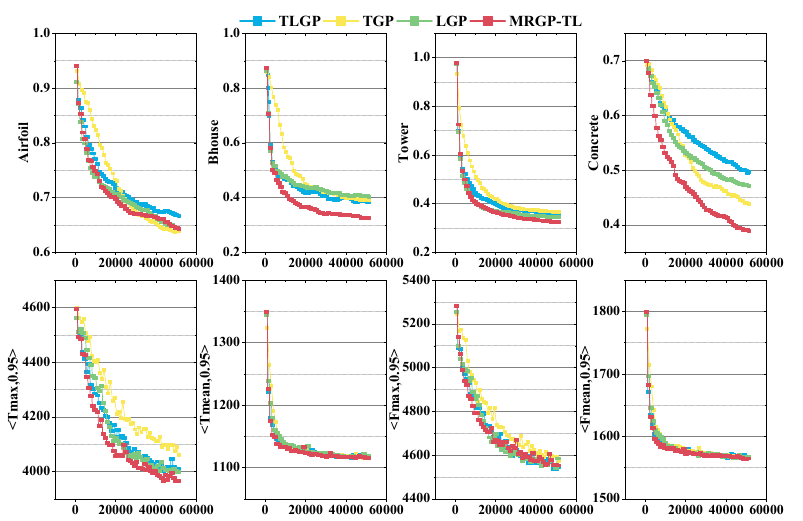}
    \caption{Test performance of the compared methods over generations in the nine symbolic regression benchmarks. X-axis: fitness evaluations. Y-axis: average test RSE for symbolic regression problems and average test objective values for DJSS problems.}
    \label{fig:SRnDJSSconv}
\end{figure}

To analyze the learning ability of the compared methods, Fig. \ref{fig:SRnDJSSconv} shows the test performance of the compared methods over generations in eight example problems. 
Specifically, we select four real-world symbolic regression benchmarks and four DJSS scenarios with a high utilization level (i.e., 0.95) as the example problems since the real-world symbolic regression benchmarks and the DJSS scenarios with a high utilization level have better real-world practical value.

MRGP-TL (i.e., red curves) has smaller test performances within fewer fitness evaluations than the others in many cases, such as BHouse and Concrete. In some other cases, though MRGP-TL levels off at a similar test performance with the other compared methods, MRGP-TL achieves the test performance earlier than the compared methods at the early stage of the evolution.
The results imply that MRGP-TL has a very competitive training performance with other compared methods in both symbolic regression and DJSS problems and can find solutions with better effectiveness within fewer simulation times in some specific cases.

\subsection{Summary}
In summary, MRGP-TL substantially improves the effectiveness of baseline methods and has a very competitive training performance with other compared methods in solving symbolic regression and DJSS problems. The results imply that sharing search information among different GP representations is a very potential direction in improving GP performance.
The experiments on tree-based and linear representations shows that MRGP-TL can automatically take advantage of the most suitable representation to achieve better performance, which is less dependent on the domain knowledge of the suitable representations for different problems than GP methods with a single representation. 
Further comparison between MRGP-TL and TLGP confirms that the performance gain of MRGP stems from the knowledge sharing among different representations, which is fulfilled by the newly proposed cross-representation adjacency list-based crossover.




\section{Further Analyses and Discussion}
To have a further understanding of the multi-representation mechanism, this section conducts six investigations based on the two applications.
First, we compare MRGP-TL with two state-of-the-art GP methods in solving symbolic regression and DJSS problems respectively to further verify the effectiveness of MRGP-TL.
Second, we analyze the average program size of the GP population over generations.
Third, we analyze the sensitivity of the key parameters in MRGP-TL. Fourth, we investigate the effectiveness of representation diversity and the effectiveness of different computation resource allocations to GP representations. Finally, we study the effectiveness of knowledge sharing between tree-based and linear representations and take two examples of adjacency lists to analyze the knowledge sharing in MRGP.

\subsection{Comparison with Latest Methods}
We compare MRGP-TL with two latest GP methods, semantic linear genetic programming \cite{Huang2022SLGP} for symbolic regression problems and grammar-guided linear genetic programming \cite{Huang2023G2LGP} for DJSS problems. These two compared methods have been published recently and have shown promising performance in symbolic regression and DJSS problems respectively. The experiment settings in this section follow the ones in \cite{Huang2022SLGP} and \cite{Huang2023G2LGP} to have a fair comparison. To further investigate the performance impact of the multi-representation mechanism, we developed two compared algorithms as well that incorporate the multi-representation mechanism with the two latest methods respectively. For the sake of simplicity, we add the latest GP (\cite{Huang2022SLGP} for symbolic regression problems and \cite{Huang2023G2LGP} for DJSS problems) as the third sub-population besides basic TGP and LGP. The three sub-populations with different representations simultaneously evolve based on the proposed mechanism. We simply denote semantic linear genetic programming and grammar-guided linear genetic programming as LATEST for symbolic regression and DJSS problems and denote both the two latest methods with the multi-representation mechanism as MRGP-latest.

\begin{table}[]
    \centering
    \caption{Average Test performance (std.) comparison among MRGP-TL and latest methods}
    \label{tb:SOTA_comp}
    \scalebox{0.75}{
    \setlength{\tabcolsep}{3pt}
    \begin{tabular}{c|c|ccc}\hline
        \multicolumn{2}{l|}{Datasets or scenarios}                                     & LATEST           & MRGP-TL         & MRGP-latest        \\ \hline
 \multirow{10}{*}{\rotatebox{90}{Symbolic Regression}} & Nguyen4                              & 0.0006 (0.001) & 0.067 (0.154) $-$ & 0.0092 (0.019) $-$ \\
                                        & Keijzer11                            & 0.0309 (0.012) & 0.289 (0.143) $-$ & 0.0474 (0.033) $\approx$ \\
                                        & R1                                   & 0.0013 (0.002) & 0.021 (0.023) $-$ & 0.0034 (0.004) $-$ \\
                                        & Airfoil                              & 0.402 (0.022)  & 0.547 (0.106) $-$ & 0.3905 (0.046) $\approx$ \\
                                        & Bhouse                               & 0.2123 (0.028) & 0.346 (0.143) $-$ & 0.223 (0.028) $-$  \\
                                        & Tower                                & 0.1367 (0.012) & 0.305 (0.037) $-$ & 0.1441 (0.016) $-$ \\
                                        & Concrete                             & 0.1871 (0.014) & 0.37 (0.085) $-$  & 0.2105 (0.027) $-$ \\
                                        & Redwine                              & 0.6514 (0.014) & 0.735 (0.034) $-$ & 0.6621 (0.025) $\approx$ \\ \cline{2-5}
                                        & mean rank                            & 1.125          & 3               & 1.875            \\
                                        & p-value(vs. SOTA)                              &                & 0.001           & 0.401            \\\hline \hline
\multirow{14}{*}{\rotatebox{90}{DJSS}}                 & $\langle$Tmax,0.85$\rangle$   & 1922.1 (42.9)  & 1926.9 (79.3) $\approx$ & 1920.3 (35.3) $\approx$  \\
                                        & $\langle$Tmax,0.95$\rangle$   & 3943.1 (84)    & 3964.2 (89) $\approx$   & 3992.4 (121) $\approx$   \\
                                        & $\langle$Tmean,0.85$\rangle$  & 417.7 (2.6)    & 417.2 (3.4) $\approx$   & 408.7 (2.7) $+$    \\
                                        & $\langle$Tmean,0.95$\rangle$  & 1116.7 (8.7)   & 1116.5 (10.6) $\approx$ & 1086.2 (12.9) $+$  \\
                                        & $\langle$WTmean,0.85$\rangle$ & 723.6 (7.5)    & 724 (5.7) $\approx$     & 712.7 (6.2) $+$    \\
                                        & $\langle$WTmean,0.95$\rangle$ & 1724.4 (26.6)  & 1723.5 (19.4) $\approx$ & 1712.8 (20.5) $+$  \\
                                        & $\langle$Fmax,0.85$\rangle$   & 2534.6 (74.1)  & 2493 (33.3) $+$   & 2485.6 (33.6) $+$  \\
                                        & $\langle$Fmax,0.95$\rangle$   & 4599.7 (80.6)  & 4553 (110.8) $+$  & 4628.7 (114.9) $\approx$ \\
                                        & $\langle$Fmean,0.85$\rangle$  & 864.6 (3.2)    & 862.8 (2.9) $+$   & 853.9 (4) $+$      \\
                                        & $\langle$Fmean,0.95$\rangle$  & 1565.3 (10.9)  & 1565.3 (10.6) $\approx$ & 1532.1 (13.1) $+$  \\
                                        & $\langle$WFmean,0.85$\rangle$ & 1701.7 (6.1)   & 1702.8 (5.6) $\approx$  & 1689.2 (5.2) $+$   \\
                                        & $\langle$WFmean,0.95$\rangle$ & 2722.8 (25.4)  & 2711 (20.5) $\approx$   & 2694.6 (18.4) $+$  \\ \cline{2-5}
                                        & mean rank                            & 2.458          & 2.208           & 1.333            \\
                                        & p-value(vs. SOTA)                               &                & 0.091           & 0.016         \\ \hline \hline
    \end{tabular}
    }
\end{table}

Table \ref{tb:SOTA_comp} compares the test performance of the compared methods where $+$, $-$, and $\approx$ denote that a certain compared method is significantly better than, worse than, or performs similarly to LATEST respectively, based on the Wilcoxon rank-sum test. Friedman's test with Bonferroni correction and Wilcoxon rank-sum test with $\alpha=0.05$ are applied to analyze the results. 

For symbolic regression problems, MRGP-TL is significantly worse than LATEST in all the datasets. Although MRGP-latest has a statistically similar overall performance to LATEST, it is less competitive than LATEST in some datasets.

For DJSS problems, MRGP-TL has a very competitive performance with LATEST. MRGP-TL has better test performance on nine of twelve scenarios than LATEST and has a better mean rank based on Friedman's test.
Cooperating MRGP with the latest GP method further improves the performance. MRGP-latest has a significantly better overall performance than LATEST and significantly outperforms LATEST on nine of twelve DJSS scenarios based on the Wilcoxon test. 

Based on the results, we conclude that MRGP is a potential GP method but needs further investigation. It gets very promising results in DJSS problems but cannot make full use of the latest GP method when solving symbolic regression problems. 
The results imply that the multi-representation mechanism needs a more effective way to coordinate different GP representations in case some less effective representations consume a large amount of computation resources.

\subsection{Program Size}
\begin{figure}
    \centering
    \includegraphics[scale=0.7, viewport=15 30 400 280, clip=true]{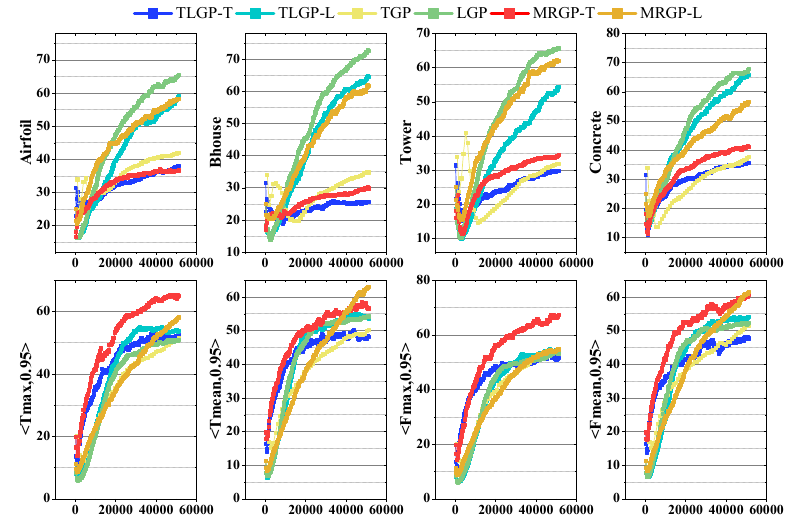}
    \caption{The average program size of the population from the compared methods over generations over 50 independent runs. X-axis: fitness evaluations, Y-axis: the average program size of the population.}
    \label{fig:programsize}
\end{figure}

To further understand the evolution of MRGP-TL, we analyze the average program size of the population in all the compared methods for solving eight example problems, as shown in Fig. \ref{fig:programsize}. Specifically, we show the program size of tree-based and linear programs respectively in TLGP and MRGP-TL, denoted by ``-T'' and ``-L'' (e.g., tree-based programs in TLGP are denoted as ``TLGP-T''). We use the tree nodes to denote the program size of tree-based programs and use the number of effective instructions multiplied by a factor of 2.0 to denote the program size of linear programs\cite{Huang2021LGP4DJSS}. We can see that the average program size from the same representation grows similarly in all the tested problems. For example, in the four symbolic regression benchmarks, TLGP-L, LGP, and MRGP-L all grow from about 20 to about 65, and TLGP-T, TGP, and MRGP-T all grow from about 10 to about 25. The similar growing curves of the same representation confirm that the proposed cross-representation adjacency list-based crossover operator has a similar variation step size with basic genetic operators and does not significantly change the average program size of the population.
Fully utilizing the interplay between tree-based and linear representations improves the effectiveness of solutions without enlarging the program size of the solutions. 

\subsection{Parameter Sensitivity Analyses}
The knowledge transfer rate among representations $\theta_t$ is a newly introduced parameter. To investigate the influence of $\theta_t$ on performance, MRGP-TL with different transfer rates are compared in this section. Specifically, we investigate the performance of MRGP-TL with a $\theta_t$ of 0\%, 10\%, 30\%, 50\%, and 70\% respectively, which are denoted as TL0 (i.e., TLGP), TL10, TL30, TL50, and TL70.

\begin{figure}[!t]
  \centering
  \includegraphics[scale=0.77, viewport=20 25 370 250, clip=true]{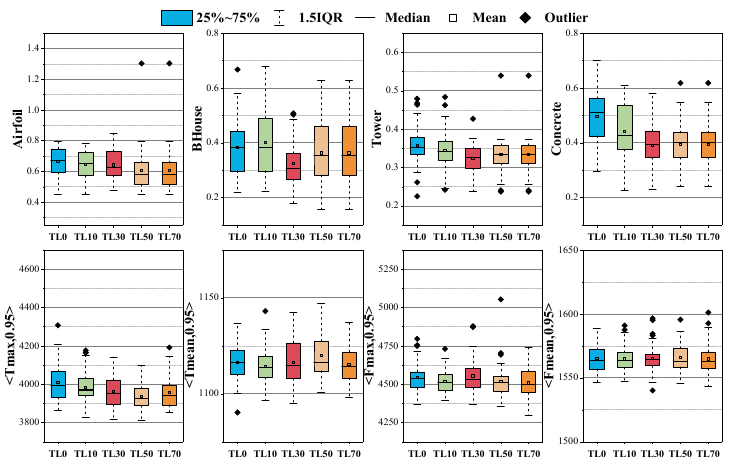}\\
  \caption{The box plots on the test performance of MRGP-TL with different $\theta_t$ values over 50 independent runs.}\label{fig:params}
\end{figure}

The test performances of MRGP-TL with different $\theta_t$s are shown in Fig. \ref{fig:params}. We see that MRGP-TL methods with $\theta_t>0$ on average have smaller (i.e., better) objective values than MRGP without any knowledge sharing (i.e., TL0 or TLGP). Besides, TL10, TL30, TL50, and TL70 have statistically similar test performance in most cases. But in some cases such as Airfoil, Concrete, $\langle T_{max},0.95 \rangle$, and $\langle F_{max},0.95 \rangle$, the increase of $\theta$ value improves the performance of MRGP-TL on average. To conclude, $\theta_t$, the knowledge transfer rate among representations shows robust performance in principle, but tuning on specific scenarios has the potential to further improve MRGP-TL performance.


\subsection{Effectiveness of CALX}
The superior performance of MRGP-TL stems from the proposed CALX operator that exchanges building blocks between the two GP representations. To further verify the effectiveness of CALX, we implement two multi-population GP methods.

Specifically, each of the two multi-population GP methods has two sub-populations using the same GP representation (i.e., tree-based or linear representations), denoted as MP-TGP and MP-LGP respectively. 
The two sub-populations exchange building blocks via basic crossover operators with the same exchanging rate (i.e., $\theta_t=30\%$).
Comparing the effectiveness of MP-T(L)GP and MRGP-TL validates the performance gain caused by the proposed crossover operator. 

\begin{table}[]
    \centering
    \caption{The average test performance (std.) of exchanging search information by basic crossover operators and CALX}
    \label{tb:mpcals}
    \setlength{\tabcolsep}{3pt}
    \scalebox{0.8}{
    \begin{tabular}{c|ccc} \hline
Datasets and scenarios               & MP-TGP          & MP-LGP          & MRGP-TL       \\ \hline
Airfoil                              & \textbf{0.582 (0.106)} $+$ & 0.648 (0.108) $\approx$ & 0.643 (0.098) \\
Bhouse                               & 0.393 (0.139) $-$ & 0.425 (0.114) $-$ &\textbf{ 0.325 (0.076)} \\
Tower                                & 0.328 (0.042) $\approx$ & 0.34 (0.043) $\approx$  & \textbf{0.325 (0.037)} \\
Concrete                             & \textbf{0.369 (0.075)} $\approx$ & 0.469 (0.107) $-$ & 0.39 (0.078)  \\
$\langle$Tmean,0.85$\rangle$  & 418 (4.8) $\approx$     & 417.2 (3.5) $\approx$   & \textbf{417.2 (3.4)}   \\
$\langle$Tmean,0.95$\rangle$  & 1118 (8.6) $\approx$    & \textbf{1115.3 (9.6)} $\approx$  & 1116.5 (10.6) \\
$\langle$WTmean,0.85$\rangle$ & 727.3 (6) $-$     & 724.4 (6.2) $\approx$   &\textbf{ 724 (5.7)}     \\
$\langle$WTmean,0.95$\rangle$ & 1737.9 (25.2) $-$ & 1724.1 (21.6) $\approx$ & \textbf{1723.5 (19.4)} \\
$\langle$WFmean,0.85$\rangle$ & 1706.5 (6.4) $-$  & \textbf{1702.4 (6.5)} $\approx$  & 1702.8 (5.6)  \\
$\langle$WFmean,0.95$\rangle$ & 2722.9 (30.2) $\approx$ & 2717.6 (23.9) $\approx$ & \textbf{2711 (20.5)}   \\\hline
mean rank                            & 2.45            & 2.2             & \textbf{1.35}          \\
p-values                             & 0.038           & 0.163           &               \\\hline
\end{tabular}
}
\end{table}

Table .\ref{tb:mpcals} shows the test performance of the three compared methods for solving ten example problems. We apply Friedman test ($\alpha=0.05$) with a Bonferroni correction to analyze the overall performance. The p-value of the Friedman test is 0.033, indicating a significant difference in the test performance of the compared methods (i.e., alternative hypothesis). The mean ranks given by the Friedman test verify that MRGP-TL has the best test performance among the three compared methods. Specifically, based on the p-values of a pair-wise comparison, we know that MRGP-TL is significantly better than MP-TGP. 

Although the performance gain of MRGP-TL is not significant compared with MP-LGP, the post-hoc Wilcoxon rank-sum test verifies that MRGP-TL has superior performance to MP-LGP on two problems and has better mean performance than MP-LGP on eight of the ten problems. The better MRGP-TL mean performance than MP-LGP in most cases leads to a further Wilcoxon test based on the ranks of the average test performance of MP-LGP and MRGP-TL. The p-value of the further Wilcoxon test is 0.005, indicating MRGP-TL is significantly better than MP-LGP in terms of average test performance. We believe that the proposed CALX effectively improves the performance of GP methods.

\begin{figure}
    \centering
    \includegraphics[scale=0.72, viewport=20 30 400 280, clip=true]{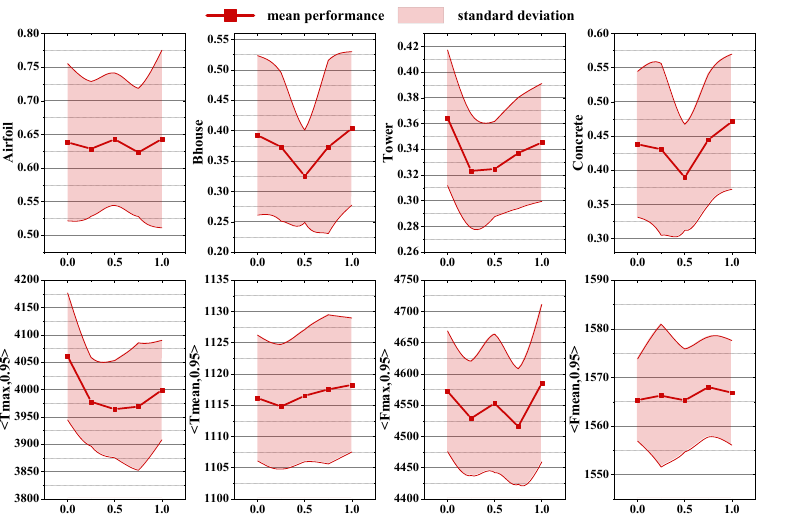}
    \caption{Test performance of different population ratios in MRGP-TL. X-axis: LGP population proportion. Y-axis: test performance of MRGP-TL}
    \label{fig:pop_rate}
\end{figure}

\subsection{Representations with Various Computation Budgets}
Different problems often have their own suitable GP representations, implying that allocating different computation resources to different representations in MRGP-TL might be helpful to the performance of MRGP-TL. To investigate the impact of computation budgets on different GP representations, we adjust the allocation of computation resources by increasing the LGP population proportion from 0\% to 100\% (and decreasing the TGP population proportion from 100\% to 0\%). Specifically, we investigate five settings of LGP proportions, which are 0\%, 25\%, 50\%, 75\%, and 100\%. The average test performance and standard deviation of MRGP-TL are shown in Fig. \ref{fig:pop_rate}.

In most of the eight tested problems, we can see a ``V'' shape roughly on the mean test performance. It implies that MRGP-TL achieves a relatively good mean test performance and standard deviation when LGP and TGP share a similar proportion of computation resources (i.e., similarly large sub-populations). Although the performance of MRGP-TL can be further improved by carefully adjusting the proportion of TGP and LGP population for a certain problem, uniformly allocating the training resources to different representations is a relatively good and robust setting for MRGP-TL.

\subsection{Benefit of Cross-representation Knowledge Sharing}
\begin{figure*}
    \centering
    \includegraphics[scale=0.77, viewport=10 25 550 240, clip=true]{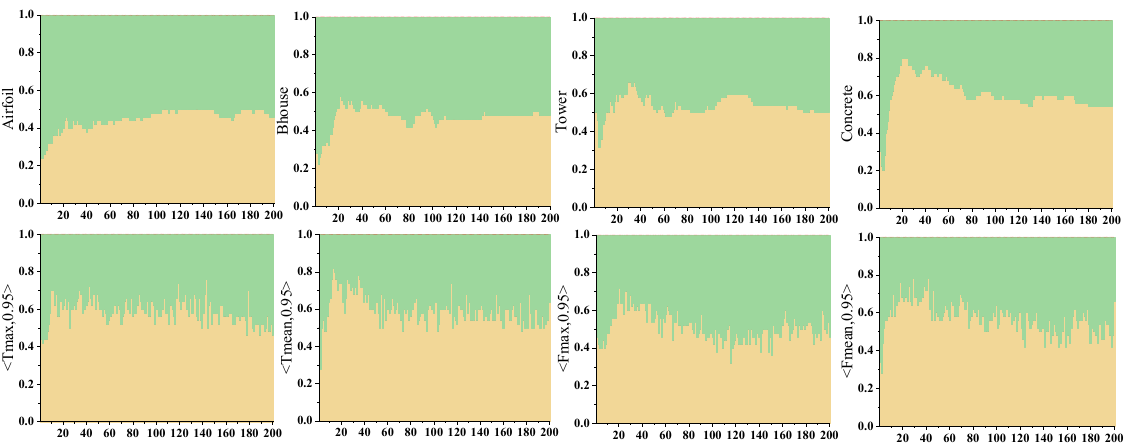}
    \caption{The average ratio of tree-based and linear representations producing the best-of-run individuals over generations in MRGP-TL. X-axis: generations, Y-axis: ratio of producing the best-of-run individuals. The green (i.e., upper) area denotes the ratio of linear representation, and the yellow (i.e., lower) area denotes the ratio of tree-based representation.}
    \label{fig:dynamics}
\end{figure*}

\begin{figure*}
    \centering
    \includegraphics[scale=0.8, viewport=10 310 530 520, clip=true]{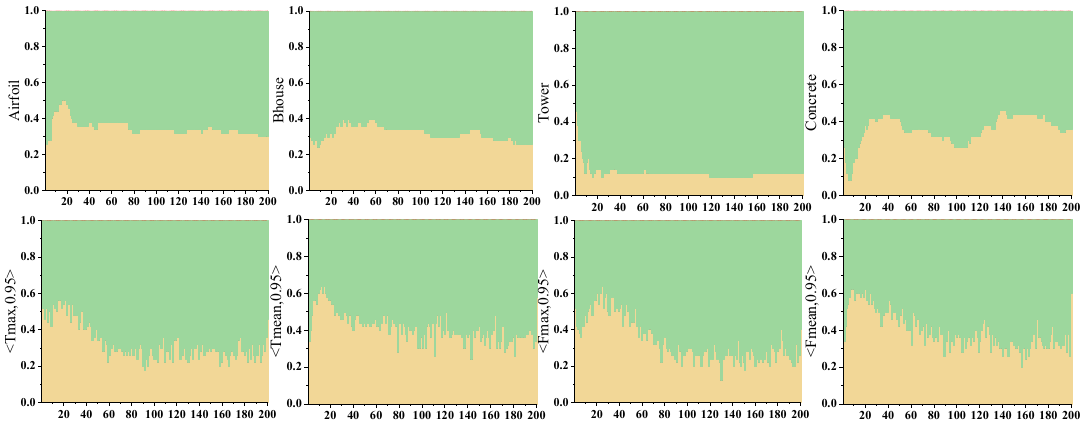}
    \caption{The average ratio of tree-based and linear representations producing the best-of-run individuals over generations in TLGP (i.e., \textbf{without knowledge sharing}). X-axis: generations, Y-axis: ratio of producing the best-of-run individuals.}
    \label{fig:dynamics_noexchage}
\end{figure*}
To verify that the knowledge sharing among representations is effective in MRGP-TL evolution, this section investigates the ratio that each GP representation produces the best-of-run individuals at every generation over 50 independent runs. 
Fig. \ref{fig:dynamics} shows the average ratio of tree-based and linear representations producing the best-of-run individuals over generations in MRGP-TL.
For comparison, Fig. \ref{fig:dynamics_noexchage} shows the ratio that different representations produce the best-of-run individuals without knowledge sharing (i.e., tree-based and linear representations in TLGP).

The two figures show that the two representations in MRGP are much less sensitive than in TLGP. In MRGP (i.e., Fig. \ref{fig:dynamics}, both tree-based and linear representations produce the best-of-run individuals with a similar ratio (i.e., 0.4$\sim$0.6) in all the selected problems. But in TLGP (i.e., Fig. \ref{fig:dynamics_noexchage}), the two representations produce the best-of-run individuals with an imbalanced ratio (e.g., linear representation produces the best-of-run individuals with nearly 90\% of runs in the course of evolution in Tower). 
It confirms that the superior representation in MRGP (e.g., linear representation in Tower benchmark) successfully improves the effectiveness of the other representation (e.g., tree-based representation), which might reduce the dependency on the domain knowledge of GP representations. Furthermore, the improvement of the inferior representation confirms that the knowledge sharing between representations effectively helps both GP representations to find more effective solutions.

MRGP effectively shares search information between representations. When a representation finds better solutions, the other representation in MRGP can be efficiently improved. For example, in the Concrete dataset, LGP has better solutions at the beginning of evolution. We can see that the tree-based representation rapidly finds more effective solutions with the help of LGP search information from generations 10 to 30. After 30 generations, effective solutions in the tree-based representation in turn help the linear representation find effective solutions and catch up with the tree-based representation at about generation 80. However, in TLGP for the Concrete dataset, the best-of-run individuals are mainly produced by the linear representation during most of the evolution (i.e., the green area covers over 60\% at each generation). Although we see a wave of the ratio between the tree-based and linear representation in the Concrete of TLGP, the wave is much smoother than MRGP since the two representations have to find effective solutions by themselves. The results confirm that knowledge sharing between tree-based and linear representations improves the performance of both tree-based and linear GP.





\subsection{Example Analyses on Adjacency Lists}
\begin{figure*}
     \centering
    \includegraphics[scale=0.65, viewport=20 20 590 190, clip=true]{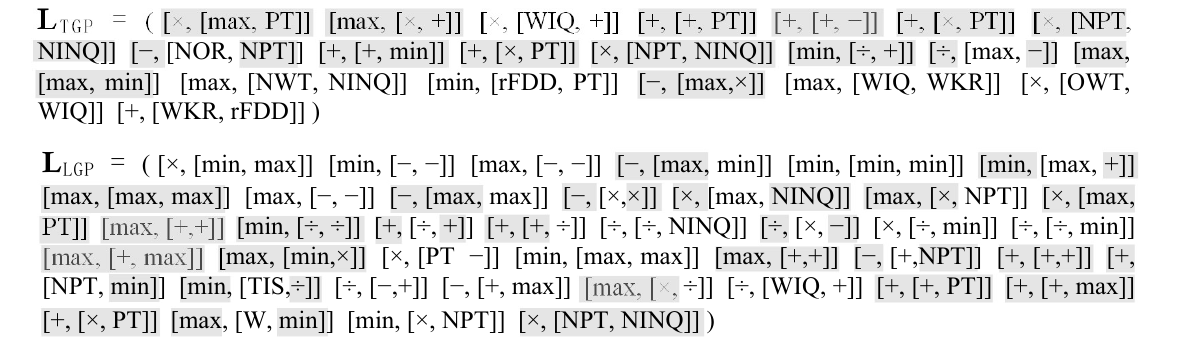}
    \caption{The adjacency lists of the outputted TGP and LGP heuristics from a run in $\langle Fmean,0.95\rangle$. The dark shadow highlights the shared adjacency of primitives between the two adjacency lists.}
    \label{fig:adjacencylist}
\end{figure*}
The proposed adjacency list-based crossover shares the search information between tree-based and linear representations by the adjacency of primitives.
To have a better understanding of knowledge sharing via adjacency lists, this section analyzes the shared knowledge (i.e., primitive adjacency) in two example adjacency lists, where each item in the adjacency lists contains two pairs of primitive connections. Fig. \ref{fig:adjacencylist} shows two adjacency lists of the best-of-run individuals from the two representations, respectively, of the same run for solving the $\langle Fmean,0.95\rangle$ DJSS problem. If a primitive connection can be seen in both of the adjacency lists, we highlight the connection with a dark shadow. For example, as the first shadowed item in $\textbf{L}_{TGP}$ shows the adjacency from ``$\times$'' to ``$max$'', the adjacency items with the same connection in $\textbf{L}_{LGP}$ are shadowed (e.g., the last item at the second line of $\textbf{L}_{LGP}$).
We can see that the adjacency lists of the output heuristics with tree-based and linear representations have a large number of shared members. For example, both of them prefer concatenating ``PT'', ``NPT'', and ``NINQ'' with ``$\times$'' and ``$+$'', which further form the shared building blocks such as ``$[+, [\times, PT]]$'' and ``$[\times, [NPT,NINQ]]$''.

Furthermore, the adjacency lists from different representations have distinct characteristics. Because of short and wide tree structures, the adjacency list of the tree-based representation considers more distinct input features, such as ``WKR'' and ``rFDD''. In contrast, the adjacency list of the linear representation uses a large number of ``max'' and ``min'' to assemble the final result. 

Overall, by exchanging adjacency lists, tree-based and linear representations can 1) learn the shared adjacency of effective solutions and 2) learn the distinct characteristics of the other representation.

\section{Conclusions}
The main goal of this paper is to verify the effectiveness of a new idea, utilizing the interplay of different GP representations to automatically identify the most suitable representation for the problem at hand.
We developed a multi-representation GP method based on tree-based and linear GP representations, denoted as MRGP-TL. Furthermore, we proposed a novel cross-representation adjacency list-based crossover operator to exchange building blocks between tree-based and linear GP representations in MRGP-TL. 
To the best of our knowledge, this paper is the first work highlighting that the interplay among different GP representations is useful for improving GP performance.

The experimental studies on symbolic regression and automatic decision rule design show that the proposed MRGP-TL significantly improves the performance of baseline GP methods without considering the interplay among different representations. Further analysis shows that MRGP-TL has a very competitive performance with state-of-the-art methods in solving DJSS problems. MRGP-TL can take advantage of a suitable GP representation in solving a certain problem, leading to a wider application spectrum.
The results also confirm that the performance gain of MRGP-TL stems from the proposed crossover operator which makes full use of the interplay between GP representations.
Fully utilizing different GP representations to enhance the search on a single task is a potential direction, which is worthy to be further investigated in other domains as well.

In the future, We will develop more effective collaboration methods among different GP representations. Adaptively and selectively evolving GP representations is a promising research direction to further improve the performance of multi-representation GP methods. We will also extend MRGP to more diverse GP representations such as gene expression programming \cite{Ferreira2001}, multi-expression programming \cite{Oltean2002}, Cartesian GP \cite{Miller1999}, and graph-based genetic programming \cite{Atkinson2018}.

\small


\end{document}